\documentclass[letterpaper, 10 pt, conference]{ieeeconf}

\IEEEoverridecommandlockouts
\usepackage{cite}
\usepackage{graphicx} % Required for inserting images
\let\labelindent\relax
\usepackage{enumitem}
\usepackage{wrapfig}
\usepackage{amsmath}
\usepackage{float}

\title{\textbf{WROOM: An Autonomous Driving Approach for Off-Road Navigation}}

\usepackage[pagewise]{lineno}
\usepackage[ruled,vlined]{algorithm2e}
\usepackage{pifont,amssymb}
\SetKwComment{Comment}{/* }{ */}
\usepackage{tabularray}

\usepackage{multirow} 
\usepackage{hyperref}
\usepackage{mathtools}

\author{
  Dvij Kalaria, Shreya Sharma, Sarthak Bhagat, Haoru Xue, John M. Dolan
}

\begin{document}

\maketitle

\begin{abstract}
    Off-road navigation is a challenging problem both at the planning level to get a smooth trajectory and at the control level to avoid flipping over, hitting obstacles, or getting stuck at a rough patch. There have been several recent works using classical approaches involving depth map prediction followed by smooth trajectory planning and using a controller to track it. We design an end-to-end reinforcement learning (RL) system for an autonomous vehicle in off-road environments using a custom-designed simulator in the Unity game engine. We warm-start the agent by imitating a rule-based controller and utilize Proximal Policy Optimization (PPO) to improve the policy based on a reward that incorporates Control Barrier Functions (CBF), facilitating the agent's ability to generalize effectively to real-world scenarios. The training involves agents concurrently undergoing domain-randomized trials in various environments. We also propose a novel simulation environment to replicate off-road driving scenarios and deploy our proposed approach on a real buggy RC car.
    Videos and additional results: \href{https://sites.google.com/view/wroom-utd/home}{\texttt{https://sites.google.com/view/wroom-utd/home}}.
\end{abstract}

% Wheeled Robot Online Off-road Mobility (WROOM)

\begin{figure}[h]
    \centering
    \includegraphics[width=1.0\linewidth]{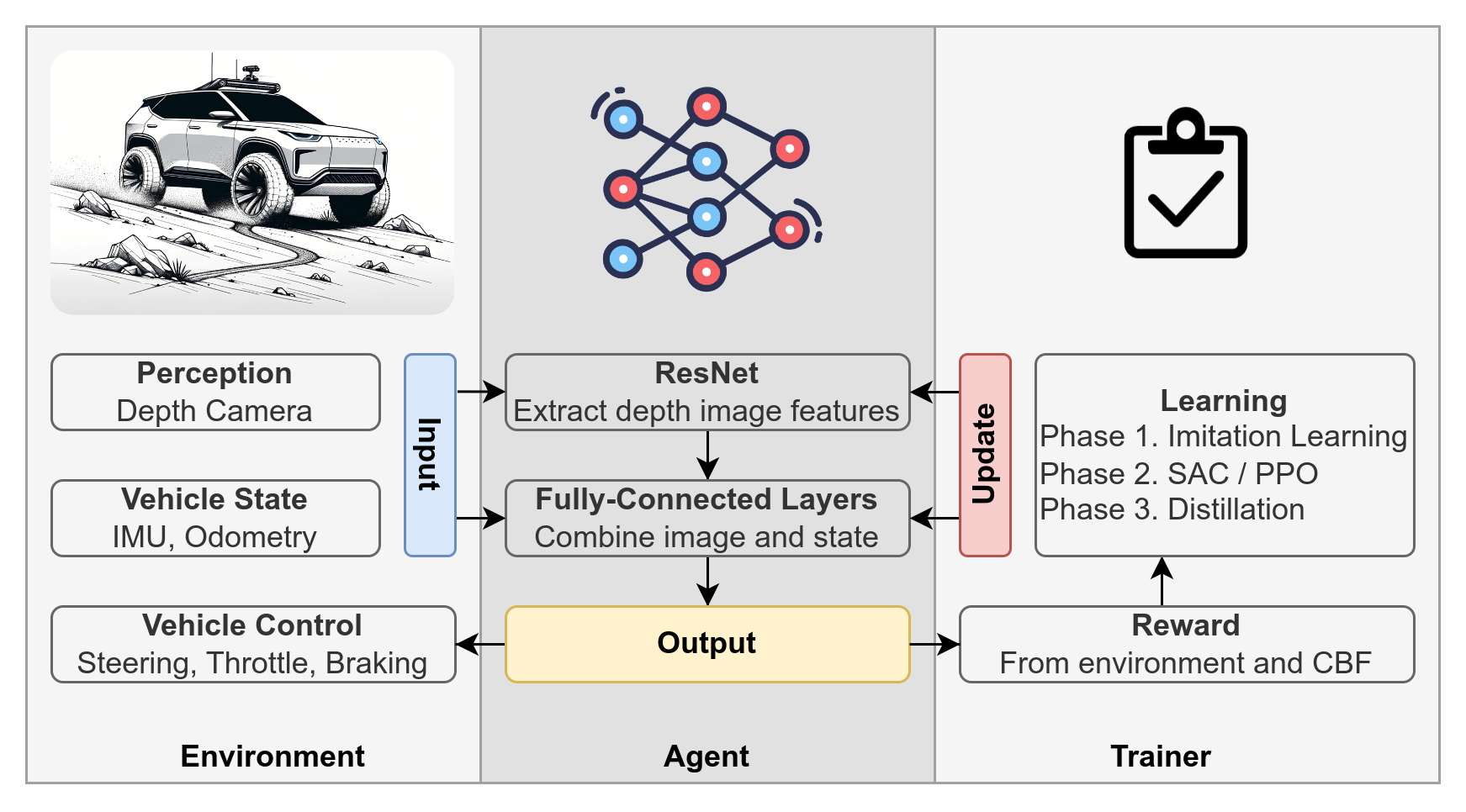}
    \caption{Overview of the proposed approach, \textbf{WROOM}: The end-to-end RL agent collects depth camera and IMU measurements from the environment and outputs steering, throttle, and braking commands. The reward function evaluates not only the agent's progress in the environment but also the smoothness and safety of its maneuvers, as reported by the control barrier function (CBF). Imitation learning is utilized to kick-start the agent, followed by Proximal Policy Optimization (PPO) to refine its policy, and finally, policy distillation is employed for real-world deployment.}
    \label{fig:intro}
\end{figure}

\section{Introduction}
% Off-road driving for a wheeled robot on a defined trail that has uneven contours. Specifically in regions with no GPS access and no prior map of the terrain. The driving should be smooth and the robot should avoid driving through terrains that could cause damage to the robot. 

Off-road driving by wheeled robots in environments characterized by uneven contours is crucial for various applications in exploration, rescue missions, and planetary exploration, which pose a significant challenge \cite{Teji2023}. Despite advancements in precise localization and state measurement technologies, traditional control systems struggle to handle the uncertain and complex dynamics introduced by challenging terrain while the vehicle is in agile motion. For instance, classic model predictive control (MPC) systems \cite{845037} encounter difficulties in coping with dynamics that lack precise and smooth mathematical descriptions. There is a strong need to address this challenge, given its relevance to real-world applications such as search and rescue missions in remote, uncharted areas, autonomous exploration in extreme environments, and operations of planetary rovers in extraterrestrial landscapes. In these contexts, conventional navigation methods reliant on GPS and pre-existing maps often prove inadequate, underscoring the need for innovative solutions.

Numerous recent studies have capitalized on classical planning strategies coupled with control design methodologies \cite{Wang_2023, datar2023wheeled, 9213856}. While some focus on generating traversability maps at the planning level, others delve into control-level techniques such as learning residual dynamics models from data \cite{Karnan2022VIIKDHA, Kalaria2023AdaptivePA}. Additionally, imitation learning-based approaches have emerged, leveraging human-collected expert data on off-road navigation environments to train agents for maneuvering through challenging terrains \cite{Pan2017AgileOA, Lu2022ImitationIN}.

While robot learning on challenging terrains has been exemplified predominantly on legged platforms \cite{agarwal2022legged, joonho2020}, there remains a conspicuous dearth of research on off-road autonomy tailored specifically for ground-wheeled robots. This scarcity underscores a notable gap, motivating the present study aimed at training a time-optimal reinforcement learning (RL) agent proficient in off-road mobility. Significantly, the task of training an end-to-end policy poses inherent challenges for wheeled robots, demanding both the ability to plan smooth trajectories and the agility to execute them effectively.

% A slightly different approach was proposed in \cite{joonho2020}, where only robot state information is used to estimate the environment to avoid common failures of exteroceptive sensors in the real world A low-level, high-frequency joint trajectory controller is also in place, making this RL system not end-to-end
% In such circumstances, it becomes imperative to develop robust and intelligent robotic systems capable of navigating autonomously solely with the onboard perception system while ensuring both the smoothness of their driving and the avoidance of paths that may pose a threat to the robot. 

% The need for addressing this challenge is underscored by real-world applications, such as search and rescue missions in remote, uncharted areas, autonomous exploration in extreme environments, and planetary rovers' operations in extraterrestrial landscapes. In these contexts, traditional navigation methods, which heavily rely on GPS and pre-existing maps, often fall short, necessitating innovative solutions.  
% Furthermore, a substantial disparity exists in the effectiveness of current Reinforcement Learning methods when applied to practical real-world scenarios beyond their simulation environment.

In pursuit of advancing off-road navigation solutions for wheeled robots in uncharted terrains, we present a novel simulation environment, \textbf{OffTerSim}, developed within the Unity game engine. This simulator serves as a training ground for agents tasked with off-road navigation, with subsequent deployment onto a real RC car. We conduct a benchmarking study employing various popular policy learning algorithms, including Control Barrier Functions (CBFs) \cite{Ames2019ControlBF}, Generative Adversarial Imitation Learning (GAIL) \cite{Ho2016GenerativeAI}, and policy distillation \cite{Rusu2015PolicyD}, within the simulator environment. Our proposed RL-based approach \textit{Wheeled Robot Online Off-road Mobility} (\textbf{WROOM}), addresses the critical need for robust off-road navigation strategies by integrating aspects of smooth driving typically handled by classical planners and robot safety usually managed at the control level through classical approaches.

In summary, our contributions can be outlined as follows:

\begin{itemize}[itemsep=0pt]
    \item We introduced a novel offroad driving simulator that procedurally generates random trails with various obstacles, enabling the training of agents capable of generalizing to real-world scenarios.
    \item We propose a novel simulation environment, \textbf{OffTerSim}, that closely replicates off-road driving scenarios that could be used for sim-to-real training of agents.
    \item To the best of our knowledge, our approach is the first one to address the challenge of offroad driving in real-world situations with sim-to-real reinforcement learning, marking a significant contribution in the field.
\end{itemize}

\section{Method}

In this section, we delve into the different components of our proposed methodology. This encompasses the design of the expert controller for executing imitation learning, and the training process for our PPO-based agent, alongside detailed descriptions of the reward functions and CBF constraints that we implement. Additionally, we touch upon the approach employed for distillation, facilitating the agent's ability to generalize effectively to real-world scenarios.

% \textbf{Problem Statement.}
% There are not many effective reinforcement learning-based adaptation methods for autonomous off-road driving for wheeled robots in unmapped environments with uneven terrains. The main reasons for this could be attributed to the lack of an existing open-source simulator for training RL agents for off-road driving settings. Also, there is a lack of standard metrics and/or environments for comparing various offline and online off-road driving algorithms.
\vspace{-1mm}
\subsection{Simulator design}
\vspace{-1mm}

% \begin{wrapfigure}{r}{0.3\textwidth}
%     \centering
%     \includegraphics[width=0.25\textwidth]{images/scandots.png}
%     \caption{Scandots (in purple) as privileged information to the expert controller.}
%     \label{fig:scandots}
% \end{wrapfigure}

\begin{figure}[h]
    \centering
    \begin{minipage}[b]{0.3\textwidth}
        \centering
        \includegraphics[width=1\textwidth]{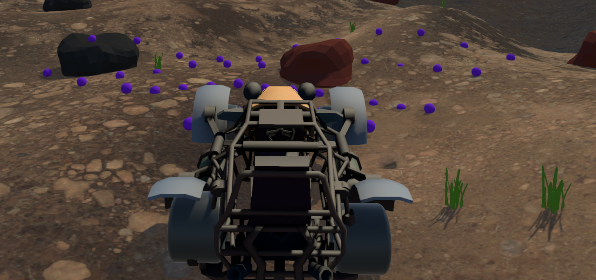}
        \caption{Scandots (in purple) as privileged information to the expert controller.}
        \label{fig:scandots}
    \end{minipage}
    \hfill
    \begin{minipage}[b]{0.16\textwidth}
        \centering
        \includegraphics[width=1\textwidth]{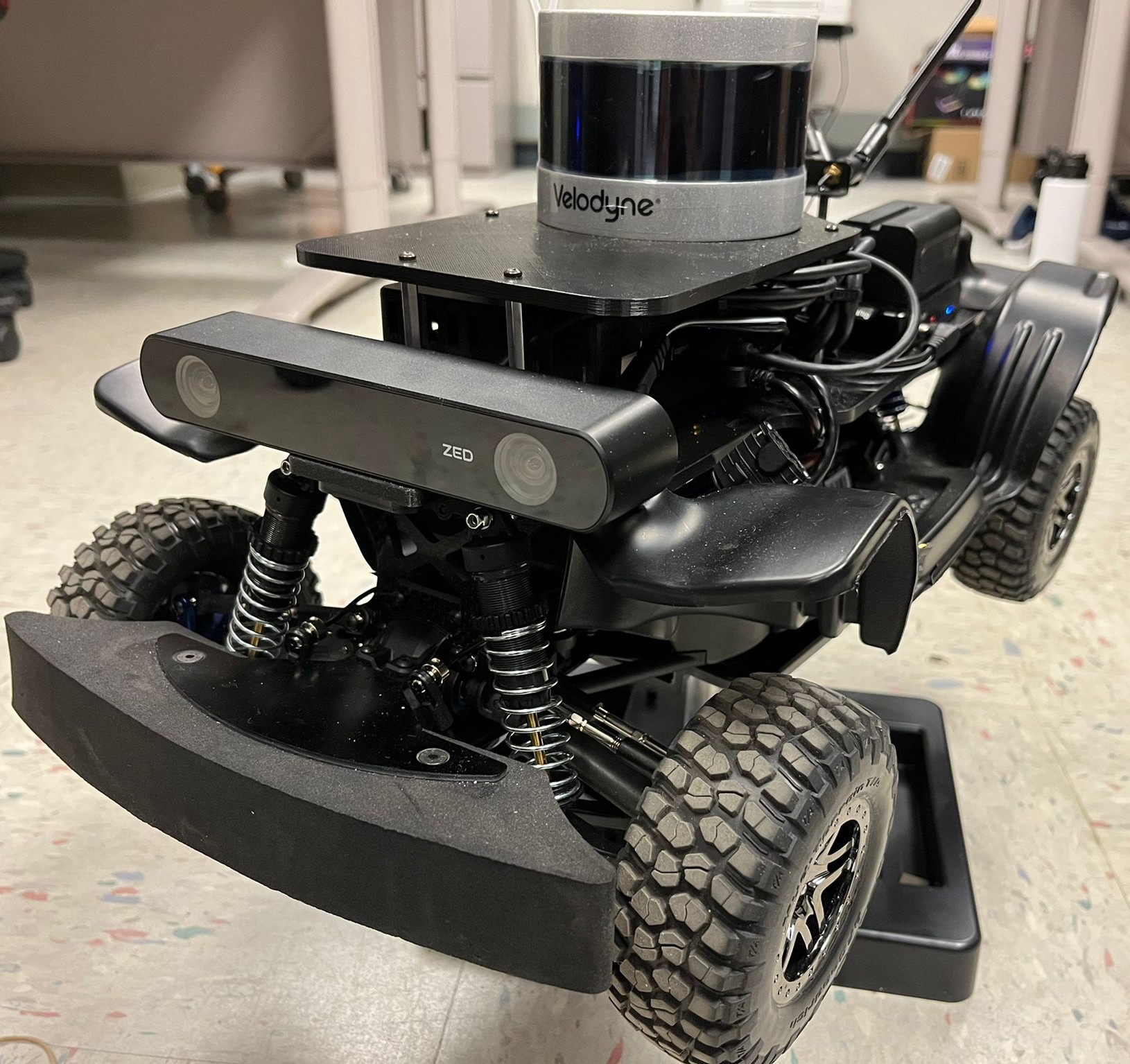}
        \caption{Real-world deployment using an RC car.}
        \label{fig:rc_car}
    \end{minipage}
\end{figure}

\begin{figure}[htbp]
  \centering
  \includegraphics[width=.5\textwidth]{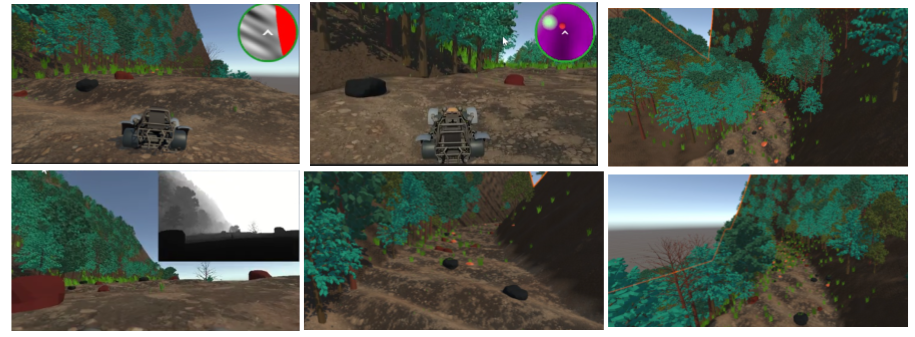}
  \caption{Stills from the proposed simulation environment, \textbf{OffTerSim}.}
  \label{fig:stills}
\end{figure}

We use the Unity Game engine \cite{haas2014history} to design a novel off-road driving simulator, which we call \textit{Off-road Terrain Simulator} (\textbf{OffTerSim}). The open-source code for the environment can be found here: \href{https://github.com/dvij542/OffTerSim}{\texttt{https://github.com/dvij542/OffTerSim}} 
We procedurally generate an environment for our simulator aiming to mimic a forest trail environment where the agent can be spawned at the start position and tasked to traverse it while taking a smooth path free of any obstacles. A random terrain is designed to mimic a forest trail. We define the centerline of the trail by a $4^\text{th}$-order equation. Without loss of generality, let us assume our origin to be the starting point of the trail where the agent will be spawned for each episode. We define our trail shape by:

\vspace{-2mm}
\begin{equation}
    y = f(x) = M/2 + b x + c x^2 + d x^3
\end{equation}
% EQN HERE
\vspace{-2mm}

where $(x,y)$ is a point on the trail, and $a,b,c,d \sim \mathcal{U}(-1,1)$ are randomly chosen coefficients that vary for each episode. The expression $w \sin \mathcal{U}(W_\text{min}, W_\text{max})$ is the width of the trail, also randomly chosen. We choose incline angles in the $x$ and $y$ directions as $\alpha_x \sim \mathcal{U}(\alpha_\text{min},\alpha_\text{max})$ and $\alpha_y \sim \mathcal{U}(\alpha_\text{min},\alpha_\text{max})$ to incline the terrain in $x$ and $y$ directions. We also bound the non-trail region on the left and right of the trail with a steepness defined by $\beta_\text{left} \text{ and } \beta_\text{right} \sim \mathcal{U}(-\beta_\text{max},\beta_\text{max})$. Based on the value of $\beta$, we can end up with hills on both sides, one side or no side to simulate all forms of trails. We also add smooth random noise to simulate unevenness on the terrain which is defined by augmenting each point on the terrain with a height value. The height map is defined by a smooth continuous $2d$ function as follows: 

\vspace{-3mm}
\begin{equation}
d(x,y) = \Sigma_{i=1..M,j=1..M} \gamma_{ij} \sin(\frac{2 \pi x}{i} + \frac{2 \pi y}{j})
\vspace{-1mm}
\end{equation}
% EQN HERE

where $\gamma_{ij}$ are Fourier coefficients chosen randomly from $\mathcal{U}(0,\gamma_\text{max})$, and $M \times M$ is the terrain image size with resolution $d$. On top of these, we also add Gaussian noise with variance $\sigma_\text{trail}$ for in-trail points and $\sigma_\text{non-trail}$ for out-of-trail points. $\sigma_\text{non-trail} > \sigma_\text{trail}$ to make the trail region smoother than the non-trail region. The vehicle model is defined as follows:

% EQN HERE
\begin{equation}
    \begin{split}
    &F_{rx} = K_\text{throttle} c_t + K_{\text{brake}} c_b  \\
    &F_{ry} = D_r \sin(C_r \arctan(B_r \alpha_r)) \\ 
    &F_{fy} = D_f \sin(C_f \arctan(B_f \alpha_f)) \\
    \end{split}
    \begin{split}
    & \hspace{-4mm} \alpha_f = c_\delta - \arctan (\frac{v_y + \omega l_f}{v_x}) \\
    & \ \ \alpha_r = \arctan (\frac{\omega l_r - v_y}{v_x})        
    \end{split}
\end{equation}

where $K_{\text{throttle}}$ and $K_\text{brake}$ are constants that dictate the longitudinal force $F_{rx}$. $l_f$ and $l_r$ are the forward and rear lengths of the vehicle from the COM. $B_f, C_f, D_f, B_r, C_r, D_r$ are the Pacejka friction coefficients of front and rear tires. $\alpha_f, \alpha_r$ are the front and rear slip angles. $\omega_z$ is the angular velocity in the body frame's $z$ axis. $v_x,v_y$ are the body frame's longitudinal and lateral velocities. $F_{ry}, F_{fy}$ are the lateral forces from the front and rear tires perpendicular to the tires. All these parameters are also varied for each episode. Unity's physics engine dictates the motion of a rigid body under the effect of all these forces. Randomly sampled obstacles of various sizes and shapes are also placed in the terrain uniformly with trees only in non-trail regions. Domain randomization includes randomly sampling all these parameters as discussed to make the agent robust to surface and trail shape unevenness. Some stills from the simulator are depicted in Figure \ref{fig:stills}.

\subsection{Reinforcement Learning (RL)}

We employ RL to train a policy that receives environmental inputs and generates actions to guide the agent along a seamless trajectory, avoiding obstacles and adhering to designated trails. This section delves into the specifics of our observation and action spaces, as well as the two different RL algorithms used for agent training.

\textbf{Observation Space.} 
Our policy model integrates various input variables, including inertial state data from the IMU sensor (acceleration, angular velocity, roll, and pitch), Frenet state \cite{5509799} (See Figure \ref{fig:state_rep}) to track lateral displacement from the center line of the trail, depth images from vehicle-mounted sensors for obstacle and path detection, and scandots \cite{agarwal2022legged} for privileged information on depth values. These inputs facilitate the agent's inference of critical information such as visible region height, Frenet frame state, lateral displacement, and heading angle from the center line.

% Our policy model relies on multiple variables serving as observations. The primary inputs encompass the inertial state obtained from the IMU sensor, including crucial parameters such as acceleration, angular velocity, as well as the roll and pitch of the vehicular agent. Additionally, we incorporate the Frenet state \ref{}, which denotes the lateral displacement from the center line of the trail. Furthermore, the model takes in as input: the depth image from the vehicle-mounted sensor. This image provides essential details about obstacles in the path and the smoothness of the visible path. Lastly, for training the privileged agent, we include an additional input (privileged information) comprising scandots \cite{agarwal2022legged}, which depict the depth values sampled at various locations in the agent's field of view. All these observation inputs enable the agent to infer information such as the height of the visible region, Frenet frame state, lateral displacement from the center line, and heading angle from the center line.
% The privileged information provided to the agent is also depicted in Figure \ref{fig:frenet}.

% \begin{wrapfigure}{r}{0.3\textwidth}
%     \centering
%     \includegraphics[width=0.3\textwidth]{frenet.png}
%     \caption{A depiction of the Frenet state frame \cite{frenet}.}
%     \label{fig:frenet}
% \end{wrapfigure}

\textbf{Action Space.} We test two action space variations for the agent. The first is continuous, offering unrestricted movement in all directions with a continuous throttle. The second variant limits heading directions to $n$ discrete values from $-1$ to $1$ with a continuous throttle control as before. The latter proves simpler for learning, as optimization becomes less challenging. Allowing continuous control in the first variant leads to the agent getting stuck due to the symmetric nature of the local obstacle avoidance problem. This results in the need to learn a discontinuous policy. In contrast, the second variant allows for easier optimization and yields qualitatively cleaner behavior.

\textbf{Policy Model Architecture.} We chose Proximal Policy Optimization (PPO) \cite{ppo} as the reinforcement learning algorithm for optimizing policies in environments with discrete or continuous action spaces. It iteratively collects data through interactions with the environment and updates the policy to maximize the expected cumulative reward. Unlike traditional methods, PPO employs a clipped surrogate objective to constrain policy updates, preventing significant deviations that could lead to instability. By balancing the exploration-exploitation trade-off with a proximal threshold, PPO continually improves the policy while ensuring stability.

To determine the agent's action based on the depth image, we use a policy model architecture featuring a ResNet \cite{He2015DeepRL} for image encoding, followed by fully connected layers combining depth image information with inputs from scandots and IMU sensor values.

\subsection{Control Barrier Functions (CBFs)}

We use the CBF (as detailed in \ref{subsec:backgroundCBF}) as a shield to guide the agent to learn a safe policy similar to \cite{Emam2021SafeRL}. We observe that the RL agent struggles at the beginning, going off-trail, which debars the agent from learning meaningful behavior later in the episode. To get rid of this, we use the CBF to refine commands from the policy to obey the safety constraints

The safe control is obtained via the following optimization program, where $K_{\text{viol}}$ is set to a high value and $u_{\text{ref}}$ is the nominal control obtained from the potentially unsafe policy:

\begin{equation} \label{eq:qp}
\begin{aligned}
\mathop{\min}\limits_{u} \quad & K_{\text{viol}} (C_{\text{right}}^2+C_{\text{left}}^2) + \|u-u_{\text{ref}}\|^2\\
    %\min_{u} &(K_{\text{viol}} (C_{\text{right}}^2+C_{\text{left}}^2) + |u-u_{\text{ref}}|^2)\\
    \text{s.t.} \quad& \quad u_\text{min} \le u \le u_\text{max} \\ 
\end{aligned}
\end{equation}

This new safe action is fed to the environment for controlling the agent. With this, the RL agent would be able to stay safe while learning, as the CBF shield defined in \eqref{eq:qp} will modify the unsafe controls from the nominal $u_\text{ref}$ to their corresponding nearest safe commands $u$ if admissible within control limits $u_\text{min}$ and $u_\text{max}$. To inform the policy update to generate safe action altogether and thus not get this override from CBF, we also add a negative reward for constraint violation proportional to the change in command caused by the CBF shield as follows: $R_{\text{constraint}} = k_{\text{constraint}} \|u-u_{\text{ref}}\|^2$.

\subsection{Reward Design}

Our reinforcement learning agent is trained using a reward function comprised of five key terms. These terms are designed to incentivize the agent's movement along trails, encourage progress, avoid obstacles, and ensure a safe traversal toward the goal. The specific components of the reward function are as follows:

\begin{itemize}[label=$\bullet$]
    \item \textbf{Progress Reward:} This term promotes the advancement of the agent along the trail by providing positive rewards for progress made.
    \item \textbf{Smoothness Reward:} We also emphasize the smoothness of the agent's trajectory by penalizing the magnitudes of pitch and roll of the vehicle.
    \item \textbf{Boundary Reward:} To maintain the agent within the trail boundaries, we penalize movement outside the designated path.
    \item \textbf{Collision Reward:} Heavy penalties are imposed to ensure that the agent actively avoids collisions with obstacles along its path.
    \item \textbf{CBF Reward:} Any violations of the CBF conditions are penalized to encourage adherence to safety constraints.
\end{itemize}

% \subsection{Our Framework}

\subsection{Policy distillation}

We observe that when directly trying to train an agent end-to-end, it takes a very long time to learn a meaningful policy on the high-dimensional depth image as observation. To address the given challenge, we implement a distillation step. Initially, we train an agent with privileged information, specifically the scandots, following previous works \cite{agarwal2022legged}. We first train an expert controller that would have privileged information for driving through rough terrain. Subsequently, we employ this expert policy trained with privileged information as the teacher and train a student policy network. The student network uses the teacher's outputs as ground truth observations. It takes depth images from the environment as input and learns to navigate relying solely on IMU values and depth images. This approach facilitates the student network in generalizing to real-world scenarios that it has not been explicitly trained on.

\section{Results}
In this section, we highlight the efficacy of the proposed approach, \textbf{WROOM}, on both the simulation environment, \textbf{OffTerSim}, as well as on the real-world deployment of a real RC car using a set of five quantitative metrics and the training average reward.

\begin{table*}
\centering
\begin{tabular}{|c|c|c|c|c|c|c|c|c|c|}
\hline
Index & Privileged & Discrete Actions & CBF & Pretrain & \# collisions & Collision time (s) & Progress & Cumulative unevenness & \# CBF Violations \\ \hline
% Original State + GAIL pre-training + discrete actions 
% & \checkmark & \checkmark & \checkmark & \textbf{15.2} & \textbf{4.68} & \textbf{595.14} & \textbf{1702.2} & \textbf{0} \\ \hline
1 & \ding{55} & \ding{51} & \ding{51} & - & 91.8 & 53.7 & 336.8 & 3095.14 & 13.0 \\ \hline
% Privileged state + continuous actions & 19.5 & 6.97 & 585.12 & 2240.6 & 2.3 \\ \hline
2 & \ding{51} & \ding{51} & \ding{51} & - & 12.4 & 5.61 & 588.4 & 2300.5 & 3.6 \\ \hline
% Privileged state + discrete actions & 12.4 & 5.61 & 588.4 & 2300.5 & 3.6 \\ \hline
3 & \ding{51} & \ding{51} & \ding{55} & - & 23.9 & 8.77 & 514.17 & 2369.3 & - \\ \hline
% Privileged state + discrete actions without CBF & 23.9 & 8.77 & 514.17 & 2369.3 & - \\ \hline
% Original state + discrete actions & 91.8 & 53.7 & 336.8 & 3095.14 & 13.0 \\ \hline
4 & \ding{51} & \ding{55} & \ding{51} & - & 19.5 & 6.97 & 585.12 & 2240.6 & 2.3 \\ \hline
5 & \ding{55} & \ding{51} & \ding{51} & DAgger & 13.5 & 5.88 & 569.2 & 2282.1 & 4.3 \\ \hline
% Original state + DAGGER + discrete actions & 13.5 & 5.88 & 569.2 & 2282.1 & 4.3 \\ \hline
6 & \ding{55} & \ding{51} & \ding{51} & GAIL & \textbf{15.2} & \textbf{4.68} & \textbf{595.14} & \textbf{1702.2} & \textbf{0} \\ \hline
\end{tabular}
\caption{Quantitative ablation of individual components of \textbf{WROOM} using various metrics.}
\label{tab:stats}
\end{table*}

\begin{figure}[htbp] 
  \centering
  \begin{minipage}[b]{0.11\textwidth}
    \includegraphics[width=1\textwidth]{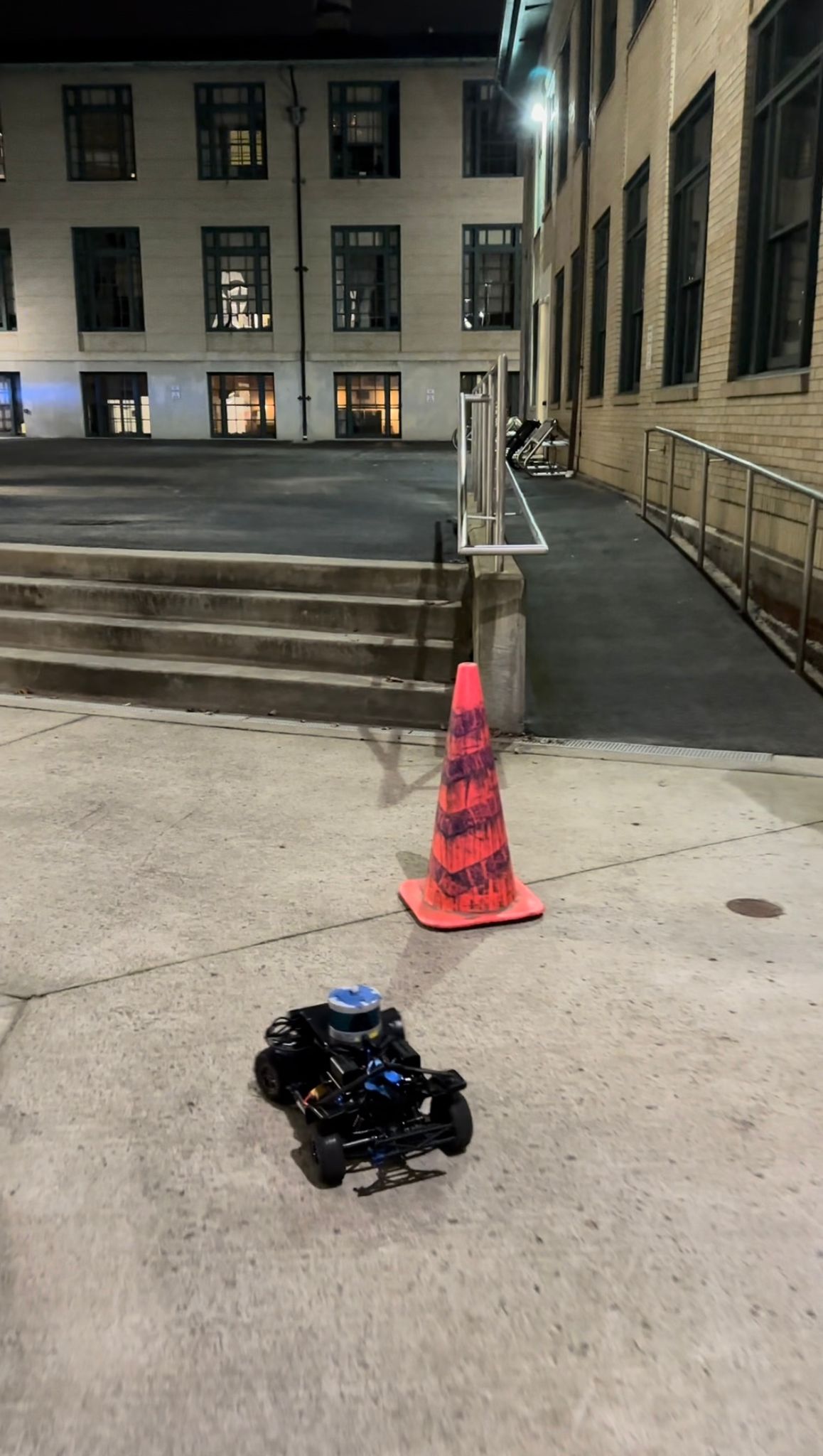}
    % \caption{Caption for Image 1}
  \end{minipage}
  \hfill
  \begin{minipage}[b]{0.11\textwidth}
    \includegraphics[width=1\textwidth]{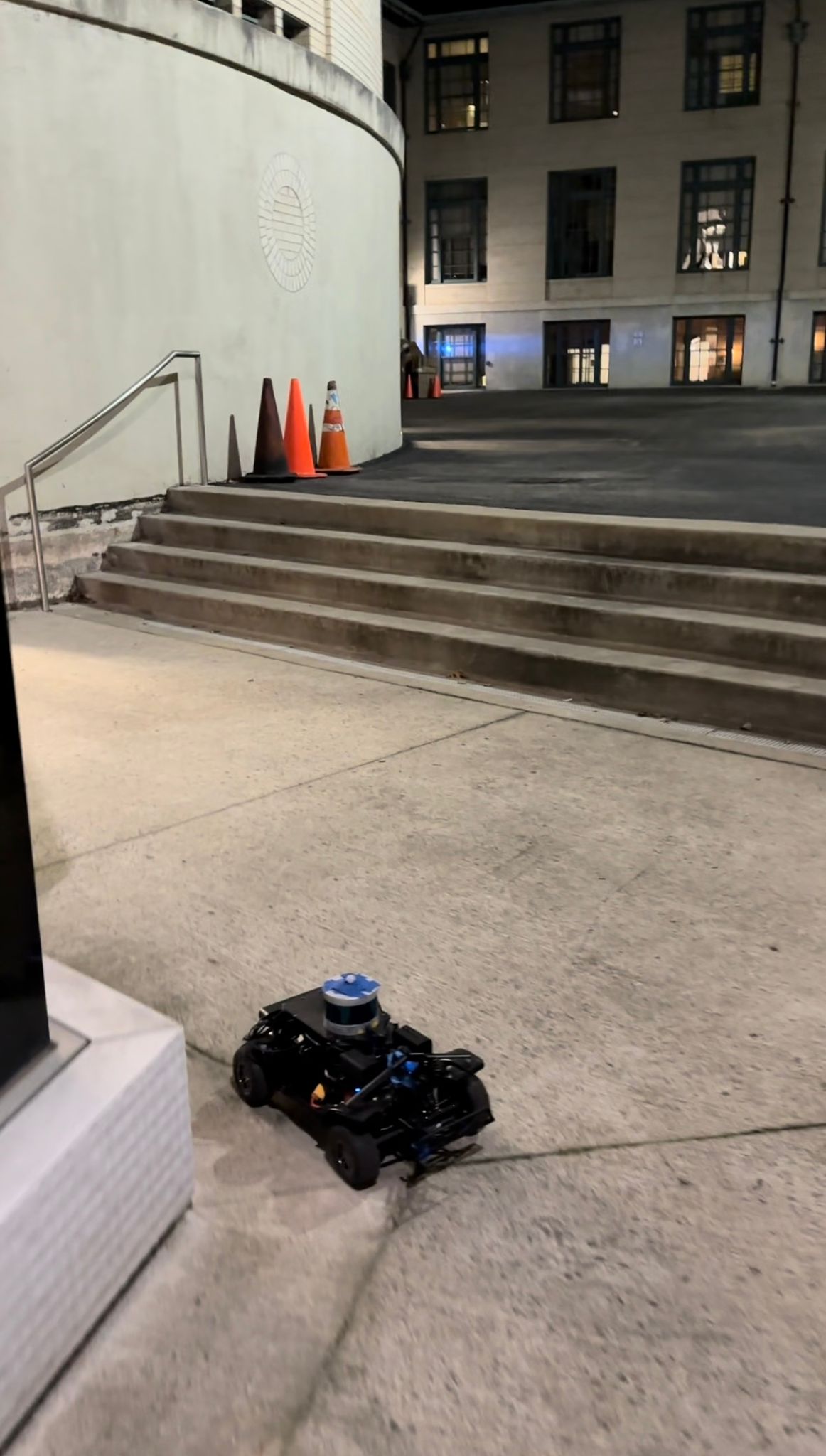}
    % \caption{Caption for Image 2}
  \end{minipage}
  \hfill
  \begin{minipage}[b]{0.11\textwidth}
    \includegraphics[width=1\textwidth]{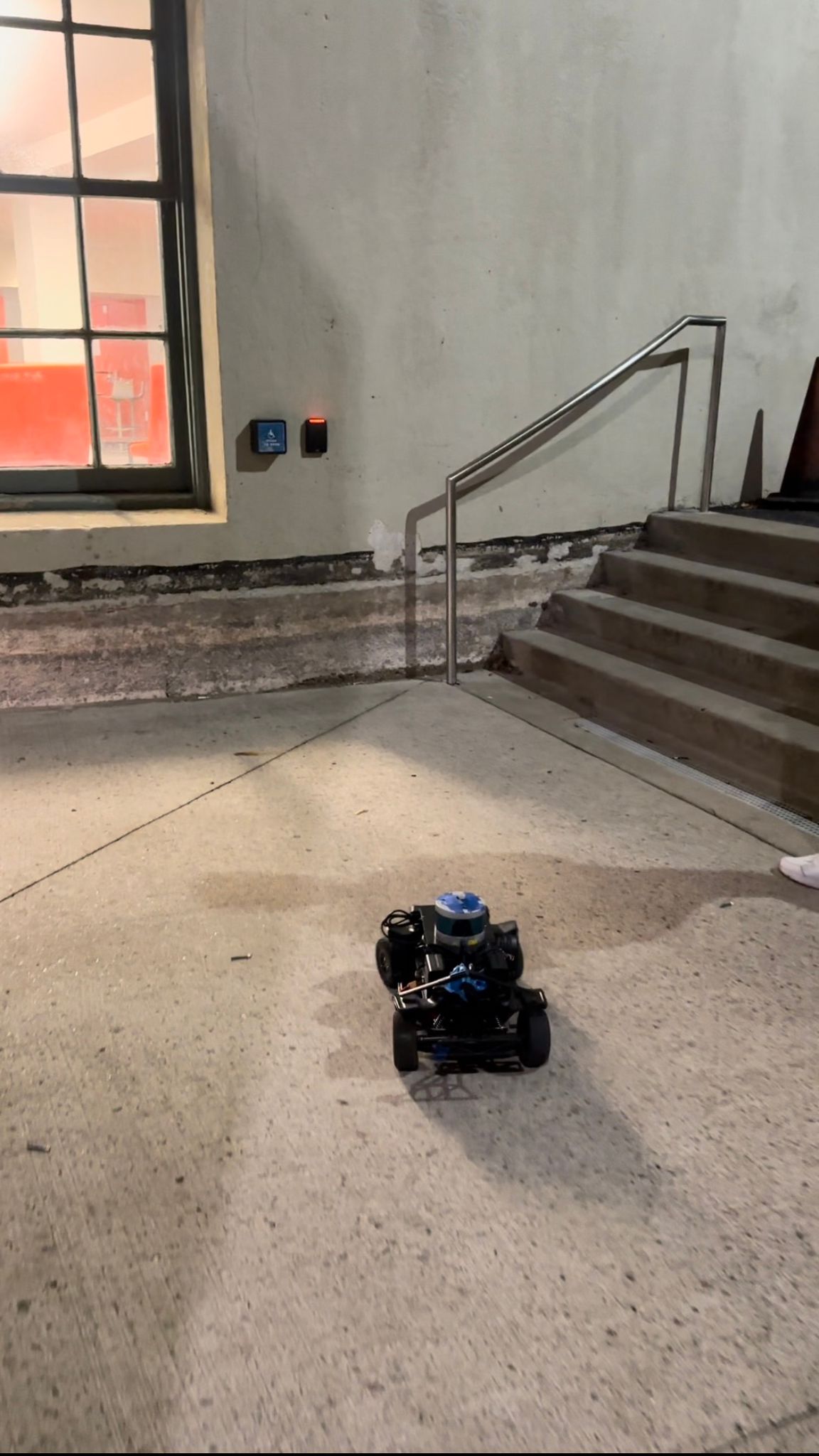}
    % \caption{Caption for Image 3}
  \end{minipage}
  \hfill
  \begin{minipage}[b]{0.11\textwidth}
    \includegraphics[width=1\textwidth]{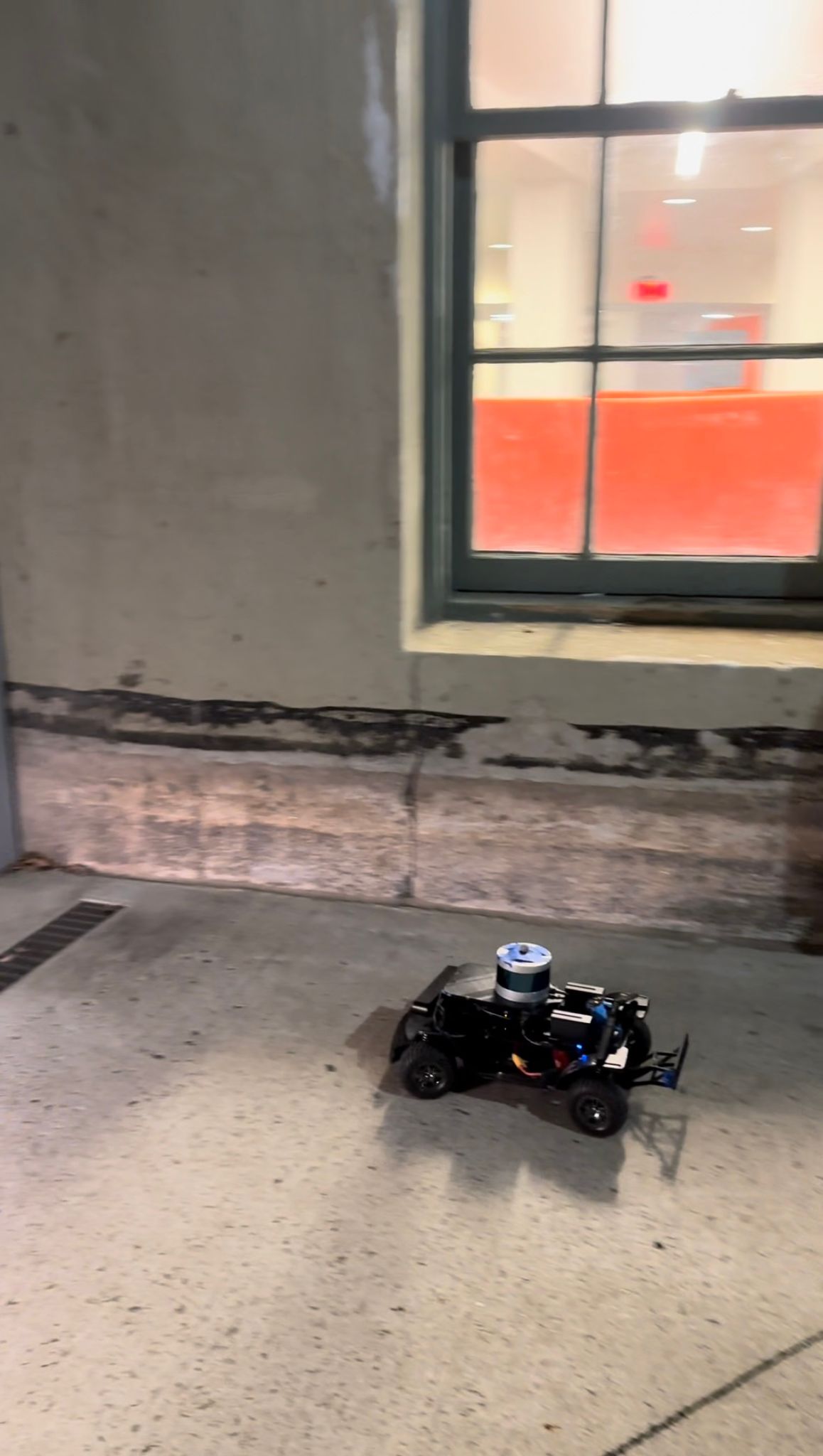}
    % \caption{Caption for Image 4}
  \end{minipage}
  \caption{Real-world deployment of our proposed approach, \textbf{WROOM}.}
  \label{fig:rc_car_example}
\end{figure}

% \vspace{-30pt}
\textbf{Metrics.} (1) \# collisions are the average number of collisions with the obstacles over 10 runs. (2) Collision time (in seconds) is the total time the car was in a collided state with the obstacle during an episode. (3) Progress is the total longitudinal distance traversed along the trail direction. (4) Cumulative unevenness (measure of rough drive)  is the sum of squares of roll and pitch over all time steps in an episode. (5) \# CBF Violations is the number of times the CBF constraints were violated in an episode.

\textbf{Quantitative Results.} We experimented with various approaches to train the agent, exploring different observations, network architectures, and output spaces. Comparative analyses of training rewards were conducted across different settings. Initially, we sought to train a policy directly using the privileged scandots information from the simulator (see Figure \ref{fig:scandots}). We employ a continuous action space for steering and throttle commands. The training involved $32$ agents concurrently undergoing domain-randomized trials in various environments.

\begin{figure}[htbp] 
    \centering
    \includegraphics[width=.9\linewidth]{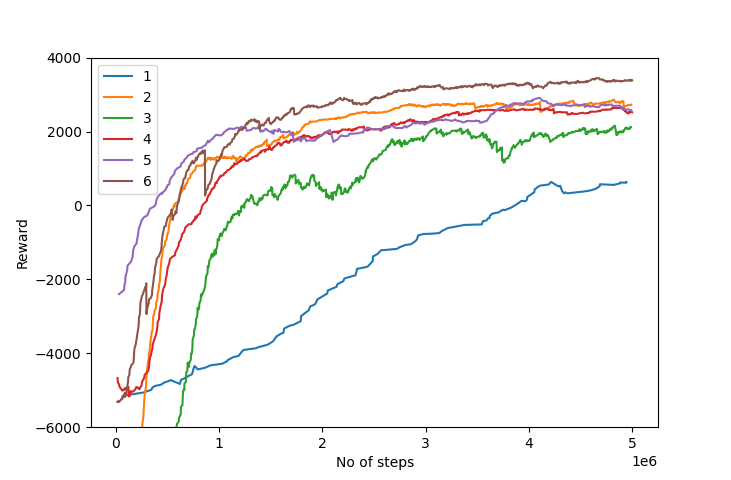}
    \caption{Training reward plots highlighting the significance of each component of the proposed approach, \textbf{WROOM}.}
    \label{fig:results}
\end{figure}

However, we encountered challenges in learning obstacle avoidance when using a continuous action space, primarily due to symmetry issues (plot 4 in Fig \ref{fig:results}). Subsequently, we transitioned to a discrete action space for steering, incorporating only $n=7$ action commands: from left to right. This adjustment resulted in significantly improved rewards, as illustrated by plot 2 in Fig \ref{fig:results}. The success of this approach can be attributed to the neural network's struggle with learning a discontinuous observation-to-action mapping when faced with obstacles directly ahead of the agent, as detailed in \cite{Almazrouei2023DynamicOA}.

% For future work, we aim to experiment with a policy trained with two outputs, dedicating one for left steering and one for right, to address the discontinuity challenge and provide a more expressive action space. However, for the outcomes presented in this work, we trained the policy with discrete steering outputs.
% We later tried to train the agent directly using only the depth image from the camera. We observed that the agent struggles a lot indirectly learning from the depth image as seen in \ref{fig:results}. 
The agent's direct training using depth-based raw observations proves ineffective due to the high dimensionality of the input, as evidenced by the examples provided in Section \ref{sec:qual} of the Appendix.
In line with the methodology outlined in \cite{agarwal2022legged}, we explored an alternative approach to distill the policy learned from privileged scandots information. Specifically, we employed DAgger-based imitation learning, as described in \cite{Ross2010ARO}, and improved metrics like \# collisions by $85.2\%$ and smoothness in drive-by $\approx 27\%$. Additionally, we experimented with GAIL \cite{Ho2016GenerativeAI}, where the expert experience was collected for each episode, and the expert controller was utilized to label the data. 
The collected data were then used to train a generator-discriminator model, imposing penalties on RL actions for substantial deviations from expert actions. The loss term scale was gradually reduced as the agent enhanced its alignment with the expert's actions. Subsequently, other RL reward terms were employed to further refine the agent's performance. 
This strategy closely resembles the policy distillation process depicted in \cite{agarwal2022legged}. 
This led to a CBF-abiding policy with no CBF violations and $82.21\%$ improvement in traversing along the set trail.

% FIG HERE (video as sequence of images)

% We endeavored to implement the trained policy on a physical RC car, depicted in Figure \ref{fig:rc_car}, utilizing specifications akin to those used in simulator training.
% \begin{wrapfigure}{r}{0.16\textwidth}
%     \centering
%     \includegraphics[width=0.15\textwidth]{images/rc_car.jpeg}
%     \caption{Real-world deployment using an RC car.}
%     \label{fig:rc_car}
% \end{wrapfigure}
\textbf{Real-world deployment.} We tested the proposed approach on a real $1/10$-scale RC car (see Fig. \ref{fig:rc_car}) on trails on the CMU campus. The car is equipped with a LiDAR, a depth dual camera, an IMU, and wheel encoders. The onboard computation platform is an NVIDIA Jetson TX2 with 8GB of RAM and a 256-core NVIDIA Pascal GPU. Some stills of \textbf{WROOM} on the real RC car can be viewed in Figure \ref{fig:rc_car_example} and the videos for the deployment can be found on the website.

 % a: Original State + GAIL pre-training + discrete actions, b: Privileged state + continuous actions, c: Privileged state + discrete actions, d: Privileged state + discrete actions without CBF, e: Original state + discrete actions, e: Original state + DAGGER + discrete actions

\section{Conclusion}

We developed a comprehensive RL system for autonomous off-road vehicles, utilizing depth camera input and allowing the model to directly output control commands. To initiate the learning process, we employed a warm start by imitating a rule-based controller. Subsequently, we utilized PPO to enhance the policy based on a reward system that integrates CBF for safety reasoning. The resulting agent demonstrates the capability to navigate challenging terrains and exhibits successful transferability to real-world scenarios. 

% more experiments on real-world trail; incoporate close-loop policy adaptation : https://arxiv.org/abs/2310.08602#:~:text=A%20critical%20goal%20of%20autonomy,limited%20to%20specific%20system%20classes.

\bibliographystyle{ieeetr}
\bibliography{main}

\begin{thebibliography}{10}

\bibitem{Teji2023}
M.~D. Teji, T.~Zou, and D.~S. Zeleke, ``A survey of off-road mobile robots: Slippage estimation, robot control, and sensing technology,'' {\em Journal of Intelligent {\&} Robotic Systems}, vol.~109, p.~38, Oct 2023.

\bibitem{845037}
J.~Rawlings, ``Tutorial overview of model predictive control,'' {\em IEEE Control Systems Magazine}, vol.~20, no.~3, pp.~38--52, 2000.

\bibitem{Wang_2023}
W.~Wang, ``Off-road driving by learning from interaction and demonstration,'' Aug 2023.

\bibitem{datar2023wheeled}
A.~Datar, C.~Pan, M.~Nazeri, and X.~Xiao, ``Toward wheeled mobility on vertically challenging terrain: Platforms, datasets, and algorithms,'' 2023.

\bibitem{9213856}
S.~Bhagat and P.~Sujit, ``Uav target tracking in urban environments using deep reinforcement learning,'' in {\em 2020 International Conference on Unmanned Aircraft Systems (ICUAS)}, pp.~694--701, 2020.

\bibitem{Karnan2022VIIKDHA}
H.~Karnan, K.~S. Sikand, P.~Atreya, S.~Rabiee, X.~Xiao, G.~Warnell, P.~Stone, and J.~Biswas, ``Vi-ikd: High-speed accurate off-road navigation using learned visual-inertial inverse kinodynamics,'' {\em 2022 IEEE/RSJ International Conference on Intelligent Robots and Systems (IROS)}, pp.~3294--3301, 2022.

\bibitem{Kalaria2023AdaptivePA}
D.~Kalaria, Q.~Lin, and J.~M. Dolan, ``Adaptive planning and control with time-varying tire models for autonomous racing using extreme learning machine,'' {\em ArXiv}, vol.~abs/2303.08235, 2023.

\bibitem{Pan2017AgileOA}
Y.~Pan, C.-A. Cheng, K.~Saigol, K.~Lee, X.~Yan, E.~A. Theodorou, and B.~Boots, ``Agile off-road autonomous driving using end-to-end deep imitation learning,'' {\em ArXiv}, vol.~abs/1709.07174, 2017.

\bibitem{Lu2022ImitationIN}
Y.~Lu, J.~Fu, G.~Tucker, X.~Pan, E.~Bronstein, B.~Roelofs, B.~Sapp, B.~A. White, A.~Faust, S.~Whiteson, D.~Anguelov, and S.~Levine, ``Imitation is not enough: Robustifying imitation with reinforcement learning for challenging driving scenarios,'' {\em 2023 IEEE/RSJ International Conference on Intelligent Robots and Systems (IROS)}, pp.~7553--7560, 2022.

\bibitem{agarwal2022legged}
A.~Agarwal, A.~Kumar, J.~Malik, and D.~Pathak, ``Legged locomotion in challenging terrains using egocentric vision,'' 2022.

\bibitem{joonho2020}
J.~Lee, J.~Hwangbo, L.~Wellhausen, V.~Koltun, and M.~Hutter, ``Learning quadrupedal locomotion over challenging terrain,'' {\em Science Robotics}, vol.~5, no.~47, p.~eabc5986, 2020.

\bibitem{Ames2019ControlBF}
A.~Ames, S.~D. Coogan, M.~Egerstedt, G.~Notomista, K.~Sreenath, and P.~Tabuada, ``Control barrier functions: Theory and applications,'' {\em 2019 18th European Control Conference (ECC)}, pp.~3420--3431, 2019.

\bibitem{Ho2016GenerativeAI}
J.~Ho and S.~Ermon, ``Generative adversarial imitation learning,'' in {\em Neural Information Processing Systems}, 2016.

\bibitem{Rusu2015PolicyD}
A.~A. Rusu, S.~G. Colmenarejo, Çaglar G{\"u}lçehre, G.~Desjardins, J.~Kirkpatrick, R.~Pascanu, V.~Mnih, K.~Kavukcuoglu, and R.~Hadsell, ``Policy distillation,'' {\em CoRR}, vol.~abs/1511.06295, 2015.

\bibitem{haas2014history}
J.~K. Haas, ``A history of the unity game engine,'' 2014.

\bibitem{5509799}
M.~Werling, J.~Ziegler, S.~Kammel, and S.~Thrun, ``Optimal trajectory generation for dynamic street scenarios in a frenét frame,'' in {\em 2010 IEEE International Conference on Robotics and Automation}, pp.~987--993, 2010.

\bibitem{ppo}
J.~Schulman, F.~Wolski, P.~Dhariwal, A.~Radford, and O.~Klimov, ``Proximal policy optimization algorithms,'' {\em ArXiv}, vol.~abs/1707.06347, 2017.

\bibitem{He2015DeepRL}
K.~He, X.~Zhang, S.~Ren, and J.~Sun, ``Deep residual learning for image recognition,'' {\em 2016 IEEE Conference on Computer Vision and Pattern Recognition (CVPR)}, pp.~770--778, 2015.

\bibitem{Emam2021SafeRL}
Y.~Emam, G.~Notomista, P.~Glotfelter, Z.~Kira, and M.~Egerstedt, ``Safe reinforcement learning using robust control barrier functions,'' {\em IEEE Robotics and Automation Letters}, no.~99, pp.~1--8, 2022.

\bibitem{Almazrouei2023DynamicOA}
K.~S. Almazrouei, I.~Kamel, and T.~Rabie, ``Dynamic obstacle avoidance and path planning through reinforcement learning,'' {\em Applied Sciences}, 2023.

\bibitem{Ross2010ARO}
S.~Ross, G.~J. Gordon, and J.~A. Bagnell, ``A reduction of imitation learning and structured prediction to no-regret online learning,'' in {\em International Conference on Artificial Intelligence and Statistics}, 2010.

\bibitem{Kalaria2023TowardsSA}
D.~Kalaria, Q.~Lin, and J.~M. Dolan, ``Towards safety assured end-to-end vision-based control for autonomous racing,'' {\em ArXiv}, vol.~abs/2303.02267, 2023.

\bibitem{xiao2023safe}
W.~Xiao, T.~He, J.~Dolan, and G.~Shi, ``Safe deep policy adaptation,'' 2023.

\end{thebibliography}

\newpage
\newpage
\appendix
\section{Appendix}
\label{sec:appendix}
\subsection{Background: Control Barrier Function}
\label{subsec:backgroundCBF}
CBFs ensure safety by rendering a forward-invariant safe set. We define a continuous and differentiable safety function 
$h(x): \mathcal{X} \xrightarrow{} \mathbb{R}$. 
The super-level set $\mathcal{C} \in \mathbb{R}^n$ can be named as a safe set. 
Let the set $\mathcal{C}$ obey
% \vspace{-10pt}
\begin{align}
    \mathcal{C} = \{\mathbf{x} \in \mathcal{X}: h(\mathbf{x})\geq 0 \} \\
    \label{eq:C}
    \partial \mathcal{C} = \{\mathbf{x} \in \mathcal{X}: h(\mathbf{x})=0 \} \\
    \text{Int}( \mathcal{C}) = \{\mathbf{x} \in \mathcal{X}: h(\mathbf{x})>0 \}.
\end{align}

A control affine system has the form $\dot{\mathbf{x}} = f(\mathbf{x})+g(x)\mathbf{u}$, such that $\exists u \quad \text{s.t.} \quad \dot{h}(\mathbf{x}) \geq -\kappa_h (h(\mathbf{x}))$, where $\kappa_h \in \mathcal{K}$ is particularly chosen as $\kappa_h(a) = \gamma a$ for a constant $\gamma > 0$. The time derivative of $h$ is expressed as $\dot{h}(\mathbf{x}) = L_fh(\mathbf{x}) + L_g h(\mathbf{x}) \mathbf{u}$, where $L_f h(\mathbf{x})$ and $L_g h(\mathbf{x})$ represent the Lie derivatives of the system denoted as $\nabla h(\mathbf{x}) f(\mathbf{x})$ and $\nabla h(\mathbf{x}) g(\mathbf{x})$, respectively. The safety constraint for a CBF is that there exists a $\gamma >0$ such that $\underset{\mathbf{u} \in \mathcal{U}}{\inf}(L_f h(\mathbf{x}) + L_g h(\mathbf{x}) \mathbf{u} ) \geq -\gamma (h(\mathbf{x}))$ for all $\mathbf{x} \in \mathcal{X}$. The solution $\mathbf{u}$ assures that the set $\mathcal{C}$ is a forward invariant, i.e.,  $x(t \xrightarrow{} \infty) \in \mathcal{C}$. 
% In a practical collision avoidance task, the safety function can be designed as the relative distance between the ego system and a dynamic obstacle.
Let us consider the state of the system as $X$. For our case, we define the state $X$ as the vehicle's state relative to the Frenet frame as $[x \ \theta \ \omega \ v \ v_\text{perp} \ c]$, where $x$ is the closest signed distance from the trail's center line with positive sign if on the right of the center line and negative otherwise. $v$ is the longitudinal velocity, $v_\text{perp}$ is the lateral velocity, $\theta$ is the relative angle wrt the center line's nearest-point tangent, and $\omega$ is the angular velocity of the vehicle. $R = \frac{1}{c}$ is the curvature of the center line at the nearest point. 
% The full state representation is shown in Fig. \ref{fig:state_rep}. 
Let $C_f$ and $C_r$ be the lateral stiffness coefficients of the front and the rear tires respectively, $l_f$ and $l_r$ are the distances of front and rear tires from the COM of the vehicle, $m$ is the mass and $I_z$ is the moment of inertia about the COM of the vehicle. 

We want to ensure that the agents stay within the trail boundaries and don't go off-trail. Hence we formulate the following CBF:

\begin{equation} \label{lane_cbf}
\small
\begin{split}
    &h_{\text{left}}(X,U) = \frac{L}{2} - x, h_{\text{right}}(X,U) = \frac{L}{2} + x\\
    &\dot{x} = v_\text{perp} \cos(\theta) + v \sin(\theta)\\
    &\dot{v}_{perp} = \frac{-2 (C_f+C_r)}{m v} v_{perp}-2 \frac{l_f C_f^2 + l_r C_r^2}{I_z v} + 2\frac{l_f C_f \delta}{I_z}\\
    &\dot{\theta} = \omega - v c \\
    &\ddot{x} = \dot{v}_{perp} \cos(\theta) + a_x \sin(\theta)+(\omega - v c) (-v_\text{perp}\sin(\theta)+v\cos(\theta))\\
    &\dot{\omega} = \frac{-2 (l_f^2 C_f + l_r^2 C_r)}{Iz v} \omega+\left(2 \frac{l_f C_f}{Iz}\right) \delta\\
    &\text{2nd-order CBF constraint for the left boundary is:}\\
    &\ddot{h}_{\text{left}}(X,U) + 2 \lambda \dot{h}_{\text{left}}(X) + \lambda^2 h_{\text{left}}(X) \ge 0 \\
    &-\ddot{x} - 2 \lambda \dot{x} - \lambda^2 \left(\frac{L}{2} - x\right) \ge 0 \\
    &\text{2nd-order CBF constraint for the right boundary is:}\\
    &\ddot{x} + 2 \lambda \dot{x} + \lambda^2 \left (\frac{L}{2} + x \right) \ge 0
\end{split}
\end{equation}

For more details about the formulation, readers are referred to \cite{Kalaria2023TowardsSA}. 
\begin{figure}
    \centering
    \includegraphics[width=0.25\textwidth]{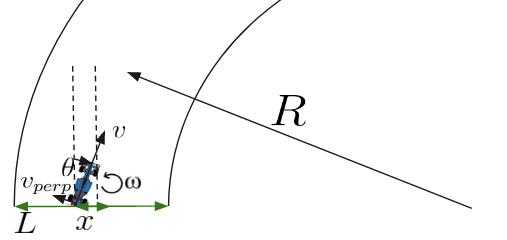}
    \caption{Vehicle state representation.}
    \label{fig:state_rep}
\end{figure}

\subsection{Implementation Details} 

To train our PPO agent, we utilize a batch size of $1024$ with a buffer size of approximately $100K$ with a learning rate of $5e^{-4}$ and a linear scheduler. The discount factor $\gamma$ is chosen as $0.99$ with a maximum horizon of $64$ time steps. The input dimensions for the depth images are chosen to be $64$.

\subsection{Qualitative Results}
\label{sec:qual}

\begin{figure}[htbp]
  \centering
  \begin{minipage}[b]{0.23\textwidth}
    \includegraphics[width=\textwidth]{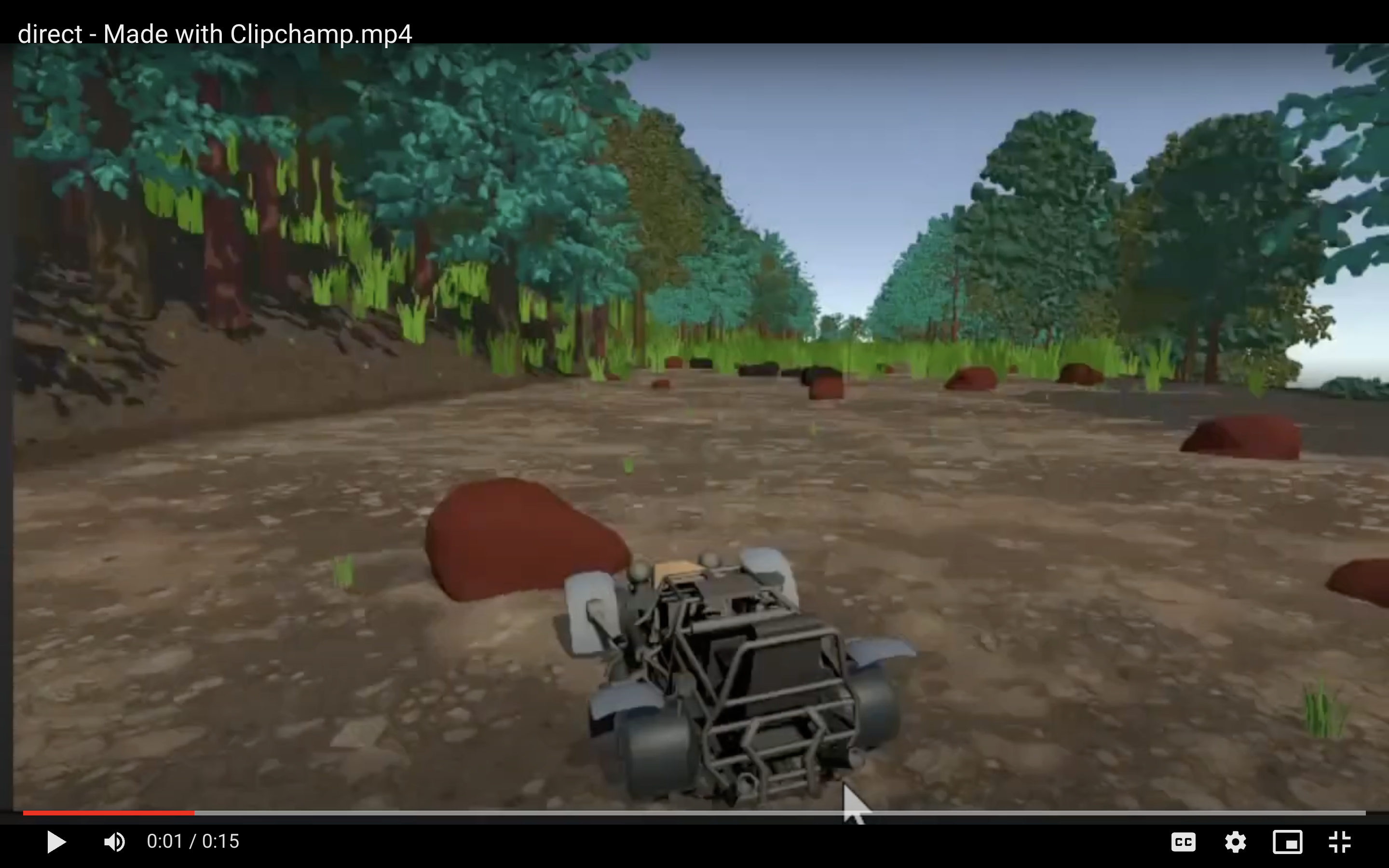}
    % \caption{Caption for Image 1}
  \end{minipage}
  \hfill
  \begin{minipage}[b]{0.23\textwidth}
    \includegraphics[width=\textwidth]{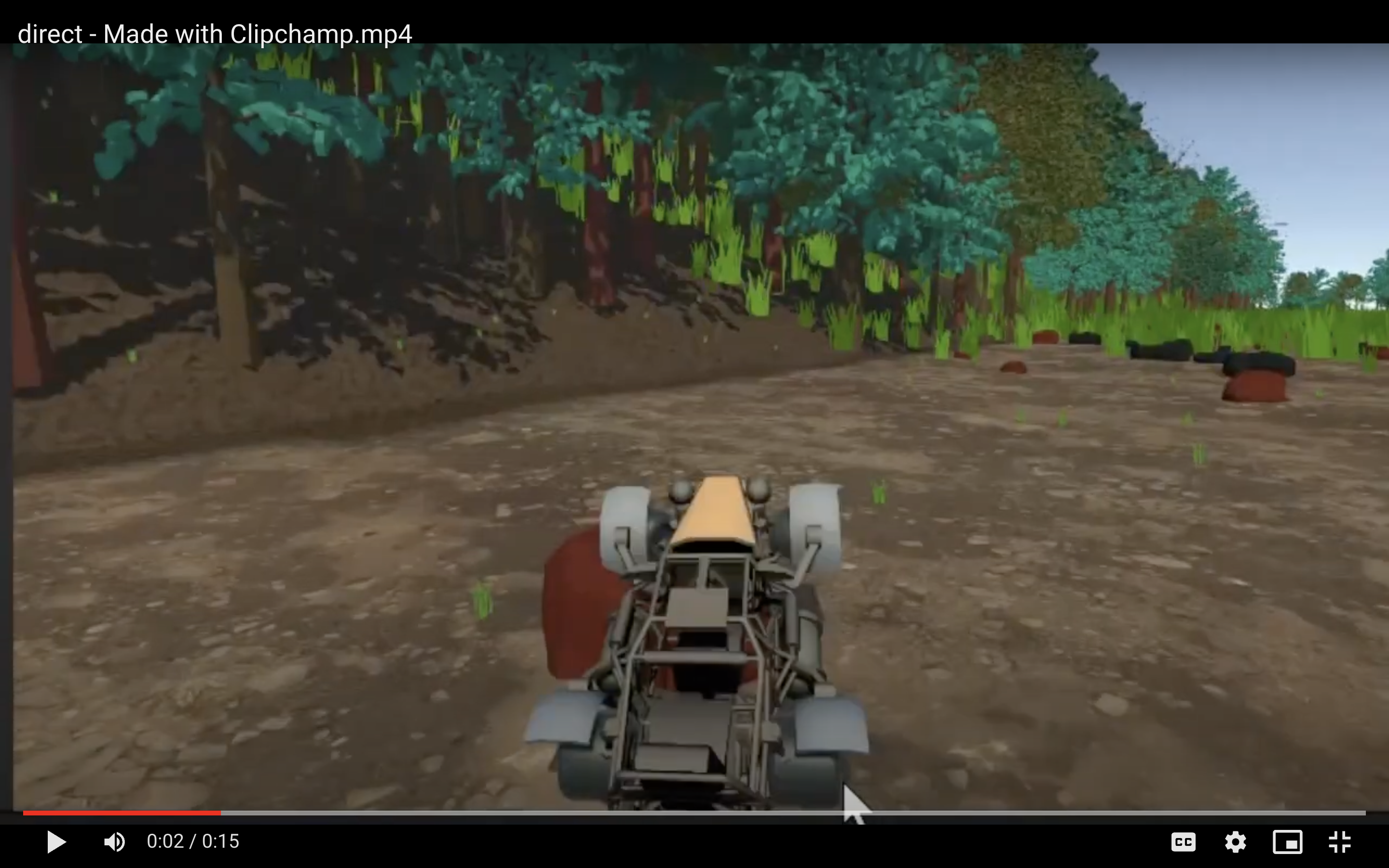}
    % \caption{Caption for Image 2}
  \end{minipage}
  \hfill
  \begin{minipage}[b]{0.23\textwidth}
    \includegraphics[width=\textwidth]{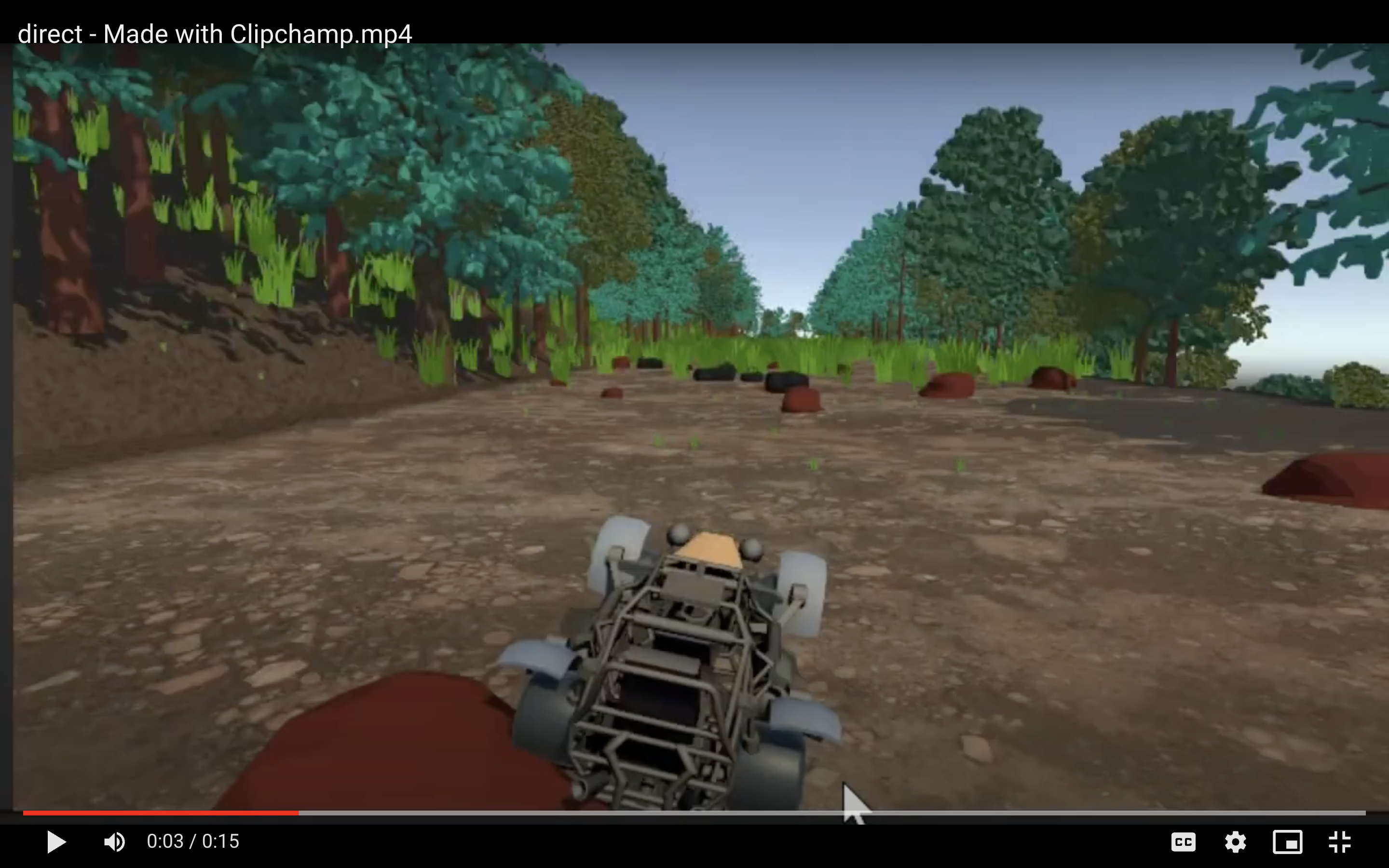}
    % \caption{Caption for Image 3}
  \end{minipage}
  \hfill
  \begin{minipage}[b]{0.23\textwidth}
    \includegraphics[width=\textwidth]{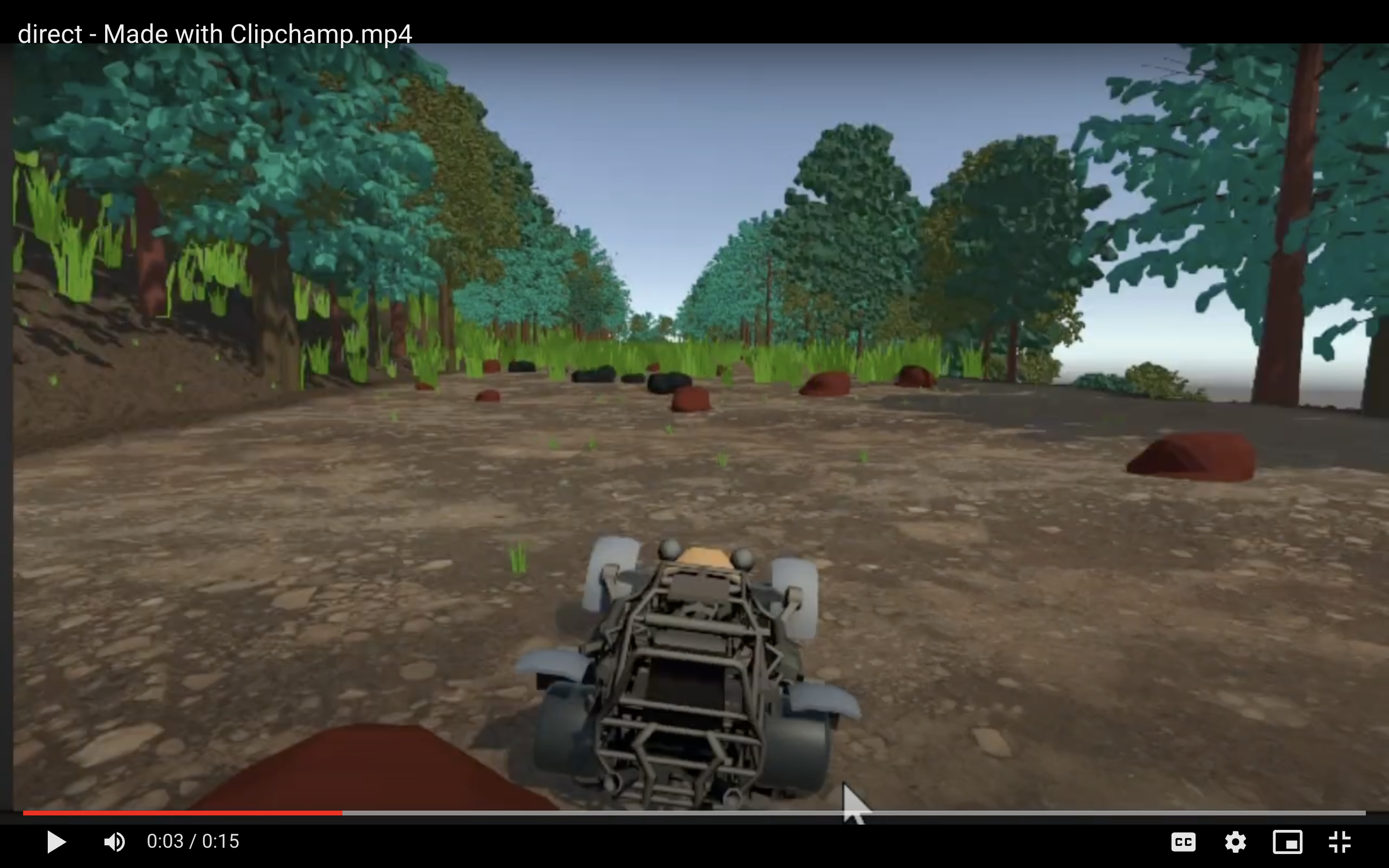}
    % \caption{Caption for Image 4}
  \end{minipage}
 \caption{Failure case when trained only on depth images}
 \label{fig:failure-sim}
\end{figure}

In this section, we present both failure case analyses and successful trial examples obtained through our approach, utilizing the proposed simulation setting. Figure \ref{fig:failure-sim} illustrates instances where our approach fails to evade obstacles in the simulator, resulting in collisions. This failure occurs because, in the depicted scenario, we solely feed depth images as raw input to the policy without access to any privileged information. Consequently, the policy struggles to discern crucial information from such high-dimensional inputs.

\begin{figure}[htbp]
  \centering
  \begin{minipage}[b]{0.23\textwidth}
    \includegraphics[width=\textwidth]{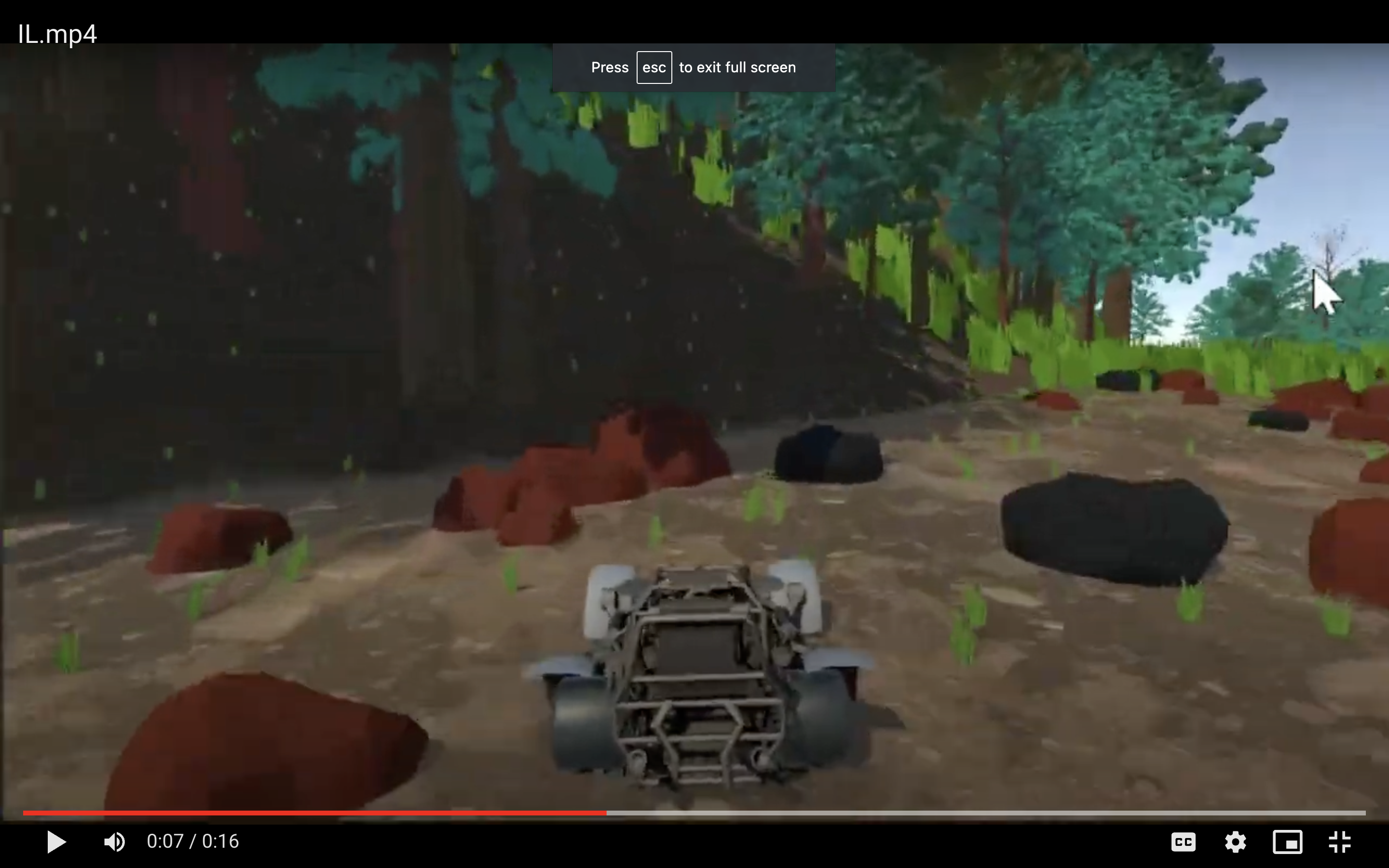}
    % \caption{Caption for Image 1}
  \end{minipage}
  \hfill
  \begin{minipage}[b]{0.23\textwidth}
    \includegraphics[width=\textwidth]{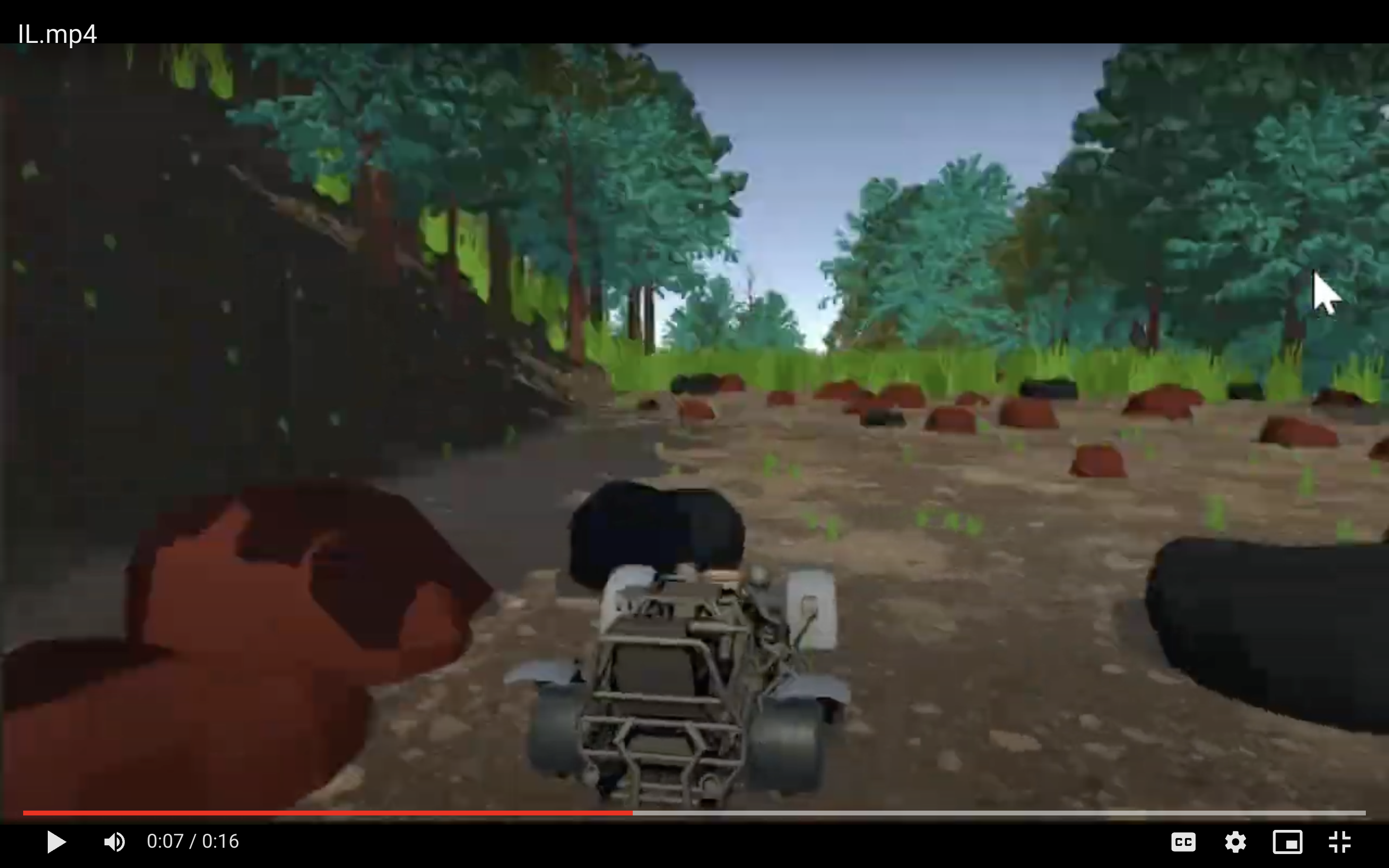}
    % \caption{Caption for Image 2}
  \end{minipage}
  \hfill
  \begin{minipage}[b]{0.23\textwidth}
    \includegraphics[width=\textwidth]{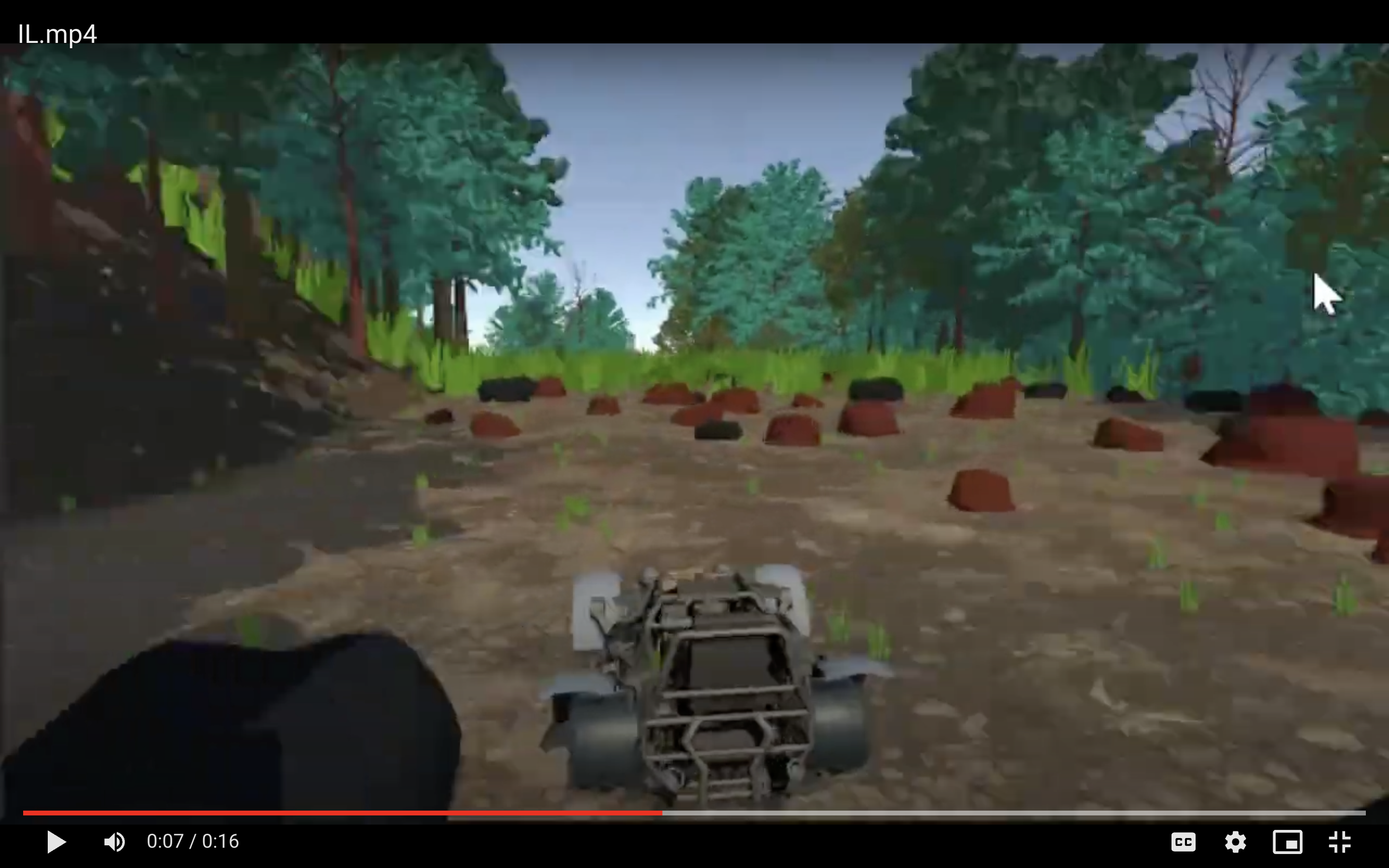}
    % \caption{Caption for Image 3}
  \end{minipage}
  \hfill
  \begin{minipage}[b]{0.23\textwidth}
    \includegraphics[width=\textwidth]{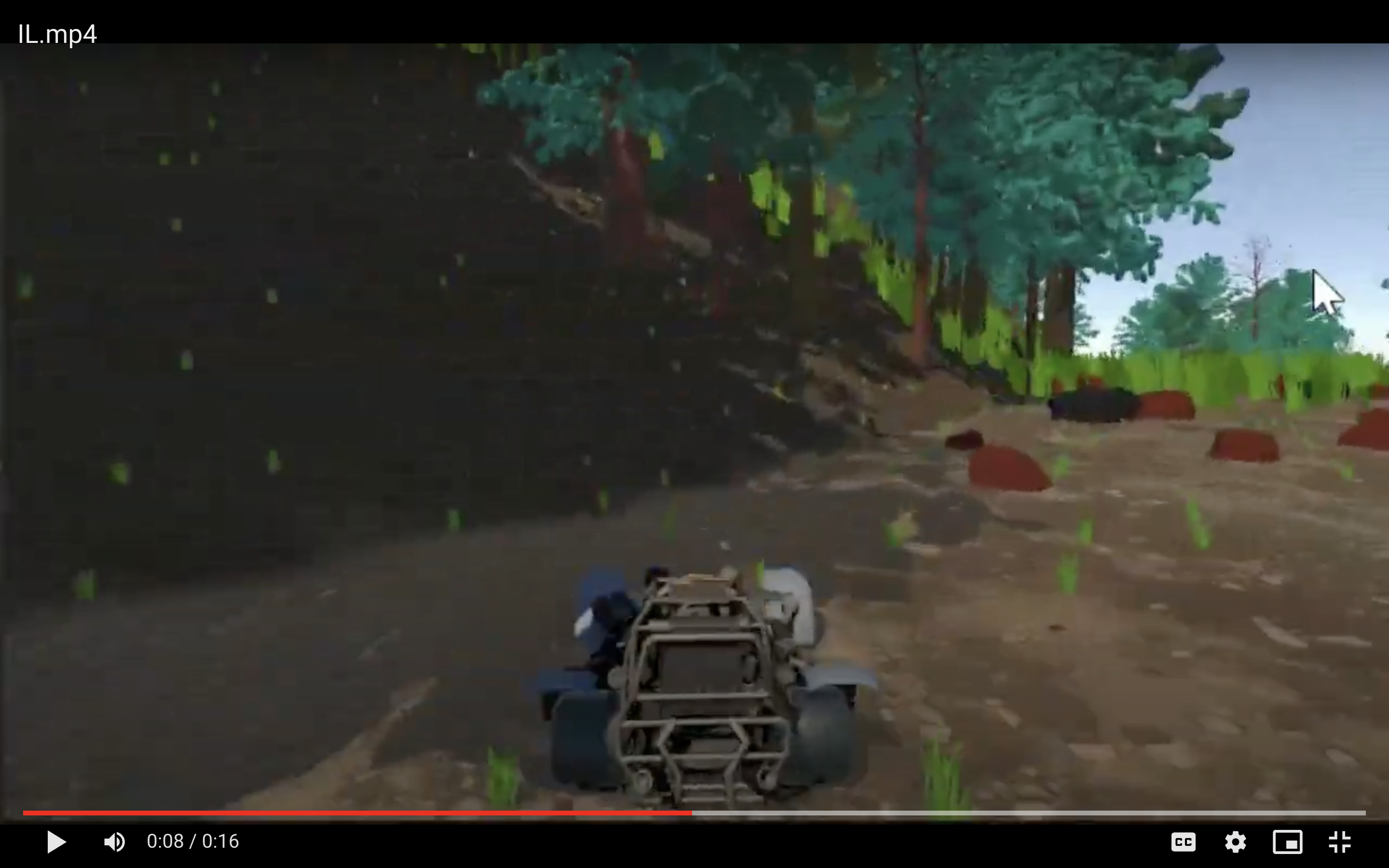}
    % \caption{Caption for Image 4}
  \end{minipage}
  \caption{Success case when using policy distillation}
  \label{fig:success-sim}
\end{figure}

However, through policy distillation, the policy acquires insights about essential information from the privileged data, which is highlighted in Figure \ref{fig:success-sim}. Consequently, when provided with depth images post-distillation, the policy comprehends how to extract vital information, even from raw depth images. This differentiation underscores the significance of the distillation pipeline in deploying our approach to real-world scenarios where depth is the only provided modality.

\subsection{Future Work}

In future endeavors, we aim to advance our RL system's robustness and adaptability by integrating closed-loop model adaptation techniques, as suggested in \cite{xiao2023safe}, to facilitate a safer and more efficient simulation-to-real transfer. Additionally, expanding our experimental scope to encompass real-world trails with diverse and intricate model parameter variations will provide valuable insights into the system's performance under more challenging conditions. Furthermore, exploring the potential integration of advanced sensor fusion techniques and leveraging state-of-the-art reinforcement learning algorithms could further enhance the system's capabilities and widen its applicability across various off-road environments and scenarios.

\end{document}